\documentclass[electronic]{vgtc}             

\ifpdf
  \pdfoutput=1\relax                   
  \usepackage{graphicx}                
  \DeclareGraphicsExtensions{.pdf,.png,.jpg,.jpeg} 
\else
  \usepackage{graphicx}                
  \DeclareGraphicsExtensions{.eps}     
\fi%

\graphicspath{{figures/}{pictures/}{images/}{./}} 

\usepackage{microtype}                 
\PassOptionsToPackage{warn}{textcomp}  
\usepackage{textcomp}                  
\usepackage{mathptmx}                  
\usepackage{times}                     
\usepackage{cite}                      
\usepackage{tabu}                      
\usepackage{booktabs}                  

\usepackage[percent]{overpic}
\usepackage{amsmath}
\usepackage{amssymb}
\PassOptionsToPackage{table}{xcolor}
\usepackage{tikz}
\usepackage{array}
\usepackage{csquotes}
\usepackage{pifont}
\usepackage{enumitem}

\vgtcinsertpkg

\onlineid{0}

\vgtccategory{Research}

\newcommand{\removed}[1]{}
\newcommand{\added}[1]{#1}

\definecolor{auto_color}{HTML}{ff7f0e}
\definecolor{truck_color}{HTML}{17becf}
\definecolor{ship_color}{HTML}{bcbd22}
\definecolor{correct_line}{HTML}{008000}
\definecolor{incorrect_line}{HTML}{ff0000}
\definecolor{correct_bar}{HTML}{D3D3D3}
\definecolor{other_bar}{HTML}{808080}
\definecolor{incorrect_bar}{HTML}{000000}

\makeatletter
\newcommand\verysmall{\@setfontsize\tiny\@vipt\@viipt}
\makeatother

\newcommand*\colorcircle[2]{\smash{\tikz[baseline=(char.base)]{
            \node[shape=circle,fill=#1,inner sep=1pt] (char) {\verysmall\bfseries\sffamily\textcolor{white}{#2}};}}}

\definecolor{yes_green}{HTML}{2CA02C}
\newcommand{\yes}{\textcolor{yes_green!90!black}{\ding{52}}}
\newcommand{\meh}{\textcolor{gray!50}{\ding{52}}}

\newcommand*\AUTOBOX{\textcolor{auto_color}{$\blacksquare$}}
\newcommand*\TRUCKBOX{\textcolor{truck_color}{$\blacksquare$}}

\newcommand{\STABLEBOX}{%
  \begingroup\normalfont
  \includegraphics[height=\fontcharht\font`\B]{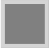}%
  \endgroup
}
\newcommand{\INBOX}{%
  \begingroup\normalfont
  \includegraphics[height=\fontcharht\font`\B]{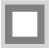}%
  \endgroup
}
\newcommand{\OUTBOX}{%
  \begingroup\normalfont
  \includegraphics[height=\fontcharht\font`\B]{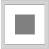}%
  \endgroup
}
\newcommand{\INOUTBOX}{%
  \begingroup\normalfont
  \includegraphics[height=\fontcharht\font`\B]{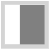}%
  \endgroup
}

\newcommand{\tasknamestyle}[1]{\textbf{\footnotesize\sffamily#1}}
\makeatletter
\newcommand{\task}[1]{%
  \tasknamestyle{#1}%
  \def\@currentlabel{\tasknamestyle{#1}}%
}
\makeatother

\makeatletter
\newcommand{\annotation}[2][gray!80]{%
  \colorcircle{#1}{#2}%
  \def\@currentlabel{\protect\colorcircle{gray!80}{#2}}%
}
\makeatother

\makeatletter
\renewcommand\paragraph{\@startsection{paragraph}{4}{0pt}%
               {.8ex \@plus 1ex \@minus.2ex}%
               {-1em}%
               {\scriptsize\bfseries\sffamily}}
\makeatother

\title{InstanceFlow: Visualizing the Evolution of\\Classifier Confusion on the Instance Level}

\author{
    Michael P{\"u}hringer%
    \thanks{E-mail: michipueh@gmail.com}\\%
    \scriptsize Johannes Kepler University Linz%
\and
    Andreas Hinterreiter%
    \thanks{E-mail: andreas.hinterreiter@jku.at or a.hinterreiter@imperial.ac.uk}\\%
    \scriptsize Johannes Kepler University Linz\\%
    \scriptsize Imperial College London%
\and
    Marc Streit%
    \thanks{E-mail: marc.streit@jku.at}\\%
    \scriptsize \centering Johannes Kepler University Linz%
}

\teaser{
  \centering
  \begin{overpic}[width=\textwidth,tics=10]{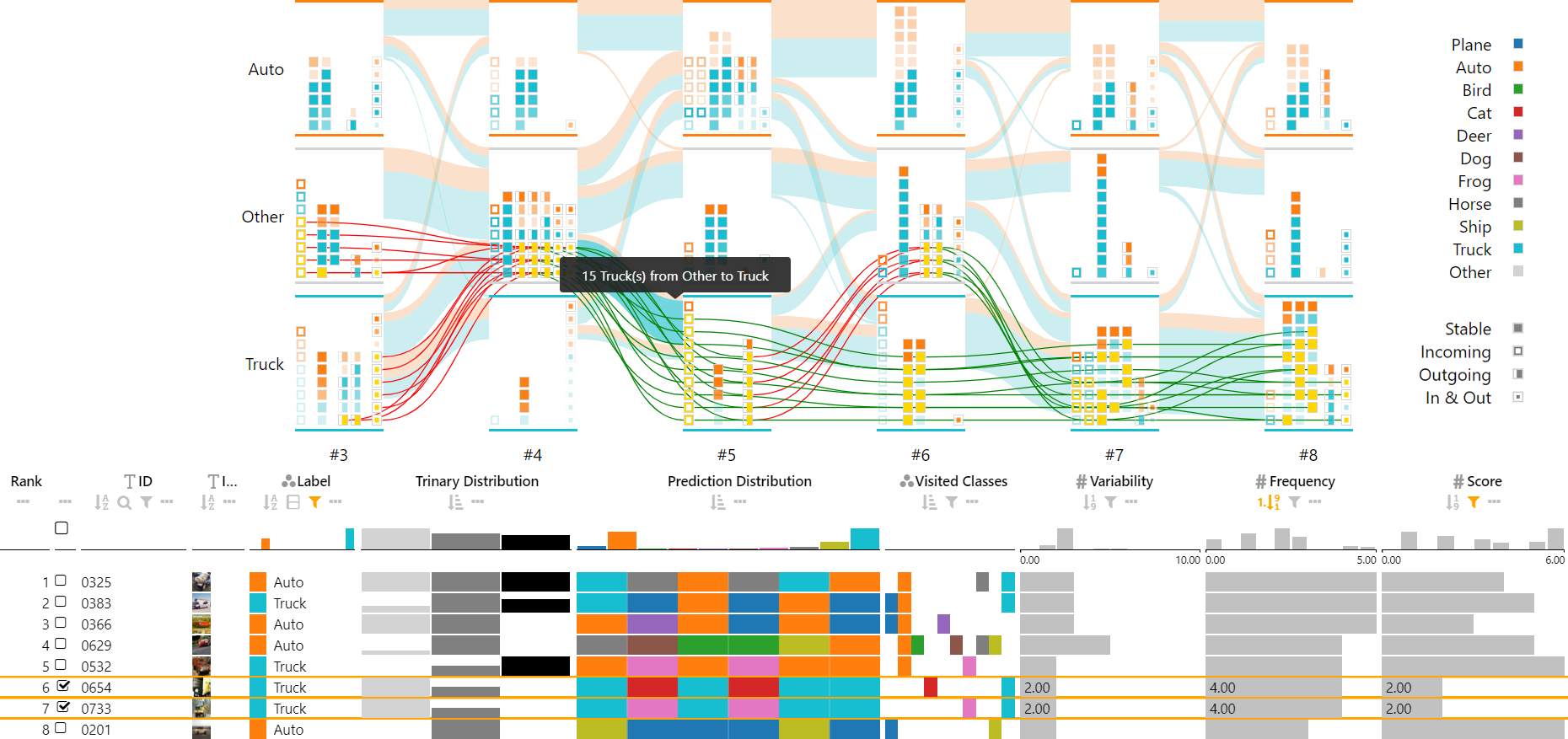}
     \put (12,37) {\annotation{A}\label{fig_label:teaser_1}}
     \put (38,15) {\annotation{B}\label{fig_label:teaser_3}}
     \put (75,15) {\annotation{C}\label{fig_label:teaser_4}}
  \end{overpic}
  \caption{InstanceFlow visualizes the evolution of a classifier's predictions throughout the training process on an instance level. The \textit{Flow View}~\protect\ref{fig_label:teaser_1} shows all instances and their corresponding class association as rectangular glyphs. A~Sankey diagram shows the fractions of instances moving between classes. Additionally, the traces of single instances can be highlighted.
  The \textit{Tabular View}~\protect\ref{fig_label:teaser_3} of the instance predictions over time along with custom performance scores~\protect\ref{fig_label:teaser_4} allows finding, ranking, and grouping instances.}
  \label{fig:teaser}
}

\abstract{Classification is one of the most important supervised machine learning tasks. During the training of a classification model, the training instances are fed to the model multiple times (during multiple epochs) in order to iteratively increase the classification performance.
The increasing complexity of models has led to a growing demand for model interpretabil\added{i}ty through visualizations.
Existing approaches mostly focus on the visual analysis of the final model performance after training and are often limited to aggregate performance measures.
In this paper we introduce InstanceFlow, a novel dual-view visualization tool that allows users to analyze the learning behavior of classifiers over time on the instance-level.
A~Sankey diagram visualizes the flow of instances throughout epochs, with on-demand detailed glyphs and traces for individual instances.
A~tabular view allows users to locate interesting instances by ranking and filtering.
In this way, InstanceFlow bridges the gap between class-level and instance-level performance evaluation while enabling users to perform a full temporal analysis of the training process.
} 


\begin{document}



\maketitle

\section{Introduction}

The real-world application of increasingly complex machine learning \added{(ML)} models has led to a growing interest in visualizations for post-hoc model explainability~\cite{arrieta_explainable_2020,hohman_visual_2018,chatzimparmpas_survey_2020}.
One of the most important supervised machine learning tasks, with a wide variety of application areas, is classification.
The performance of classification models can be analyzed and visualized on three levels of detail~\cite{confusionflow}: global\added{ly}, \added{on the} class level, and \added{on the} instance level.
Additionally, extending the analysis to cover the whole training process (multiple training iterations, i.e. epochs) has been identified as a promising research direction~\cite{liu_towards_2017,hohman_visual_2018,chatzimparmpas_survey_2020}.
However, existing approaches often focus on fully trained models and disregard\removed{ing} the temporal evolution that led to this final model state.
Furthermore, tools that enable temporal performance analysis are typically limited to global, single-value performance measures~\cite{ComparisonOfPerformanceMeasuresForClassification}.

In previous work, we argue that extending a temporal performance analysis to the class-level can lead to new insights~\cite{confusionflow}.
\removed{Still, the aggregated nature of class-level performance measures showed that a full analysis of a classification model's learning behavior also requires drilling down to the level of individual instances.}\added{Similarly, a temporal drill-down to the instance level can help model developers to distinguish stable (mis)classification patterns from stochastic effects caused by the partially random training mechanism.}

\removed{To address this issue, we introduce}\added{The main contribution of this work is} InstanceFlow, a novel visualization that combines aggregated temporal information in a Sankey diagram with detailed traces of individually selected instances.
These interesting instances can be located via a tabular view that allows users to rank and filter instances by several temporal difficulty measures.
With this dual approach, InstanceFlow aims at bridging the gap between class- and instance-level analysis of the learning behaviors of classification models.

\section{User Tasks}
\label{sec:tasks}

\removed{As stated in the introduction, }InstanceFlow focuses on a temporal analysis of instance-level classification performance.
Such an analysis can lean either towards exploring instance-based properties of certain \emph{epochs}, or analyzing the temporal characteristics of individual \emph{instances}.
Consequently, we \removed{have }structur\removed{ed} the user tasks that we seek to address with InstanceFlow by whether they are epoch- or instance-focused (see Table~\ref{tab:proposed_tasks}).

We base\removed{d} the individual user tasks on a survey of existing instance-level visualizations (Section~\ref{sec:related-work})\added{---}with a focus on filling gaps related to model-agnosticism and temporal analysis\added{---as well as on discussions with ML researchers in our previous work~\cite{confusionflow}.}
\added{InstanceFlow primarily addresses model developers and builders (cf.~\emph{Who?} in Hohman et al.~\cite{hohman_visual_2018}), who seek to better understand the training process.}

The instance-focused tasks~(\tasknamestyle{IT}) are concerned with finding instances which are hard to classify correctly~(\ref{task:inst-diff})\removed{,} or whose predictions evolve unusually~(\ref{task:inst-path}--\ref{task:inst-oscil}).
This allows users to assess temporally (un)stable \removed{weaknesses}\added{characteristics} of the model or detect \added{potentially mislabeled} input data\removed{ with potentially wrong ground truth labels}.
The epoch-focused tasks~(\tasknamestyle{ET}) are related to analyzing epoch-wise class distributions~\added{(}\ref{task:epoch-dist}\added{)} or locating problematic epochs~(\ref{task:epoch-wrong},~\ref{task:epoch-move}).
Problematic epochs are those for which weight or parameter changes produce a non-beneficial outcome, such as an increase of confusion between two critical classes.

\begin{table}[ht]
    \centering
    \caption{User tasks addressed by InstanceFlow, categorized by their focus on epoch-~(\textbf{ET}) or instance-driven~(\textbf{IT}) analysis.}
    \label{tab:proposed_tasks}
    \centering
    \footnotesize\sffamily
    \begin{tabular}{c@{~~}>{\raggedright\arraybackslash}p{.83\columnwidth}}
    \toprule
        \textbf{Task} & \textbf{Description}  \\
        \midrule
        \task{IT1}\label{task:inst-diff} 
            & Find \emph{difficult} instances \\
        \task{IT2}\label{task:inst-path}
            & Trace an instance's \removed{\emph{path} over multiple epochs}\added{\emph{classification history}} \\
        \task{IT3}\label{task:inst-visit}
            & Analyze if an instance \emph{visits many or few} classes \\
        \task{IT4}\label{task:inst-oscil}
            & Find instances \emph{oscillating} between classes \\
        \midrule
        \task{ET1}\label{task:epoch-dist}
            & Assess \emph{class distributions} for a given epoch \\
        \task{ET2}\label{task:epoch-wrong}
            & Find momentarily \emph{wrong\removed{ly} and/or correct\removed{ly}} instances\\
        \task{ET3}\label{task:epoch-move}
            & Find instances \emph{staying} in their class or \emph{moving} between classes at a given epoch \\
        \bottomrule
    \end{tabular}
\end{table}

\section{Related Work}
\label{sec:related-work}

Previous work on visualizing instance-level information in machine learning has mostly focused on model-dependent parameters such as the activation of neurons in deep neural networks in response to a given input instance.
Often, the visualizations \removed{are tailored to exploring}\added{concentrate on} the behavior of individual layers of the networks~\cite{chung_revacnn:_2016,pezzotti_deepeyes:_2018,zhong_evolutionary_2017,kahng_activis:_2018}.
A~number of works have focused particularly on the convolutional layers of CNNs~\cite{liu_deeptracker:_2018,bruckner_ml-o-scope:_2014,zeng_cnncomparator:_2017}.
Similar visualizations exist for GANs~\cite{wang_ganviz:_2018} and Deep Q-networks~\cite{wang_dqnviz:_2019}.
Most of these model-specific approaches are further limited in that they only provide information for a single \added{training} iteration at a time.

Likewise, visualizations focused on the performance analysis of classifiers typically do not enable a true temporal analysis.
Chae \removed{a}\added{e}t al.~\cite{chae_visualization_2017} show instance-wise predictions and aggregated distributions; Alsallakh et al.~\cite{alsallakh_convolutional_2018} focus on class-confusion with basic drill-down functionality to explore problematic instances.
In both cases, limited temporality is achieved via single-epoch selection sliders.

Squares by Ren et al.~\cite{ren_squares:_2017} is closely related to our work in terms of visual design and the type of information shown.
Users can switch between aggregated prediction distributions and a fine-grained instance-wise visualization using rectangular glyphs.
However, in Squares only the final model predictions can be explored.

InstanceFlow aims at enabling a true temporal performance analysis on the instance level.
In this regard, it is closely related to our previous work, ConfusionFlow~\cite{confusionflow}, which enables temporal class-level analysis of classification models via a novel adaptation of the confusion matrix.

Visually, InstanceFlow combines a multiform Sankey diagram similar to VisBricks~\cite{lex_visbricks_2011} and StratomeX~\cite{lex_stratomex_2012} with a sortable, aggreg\added{at}able tabular view (cf.~Table Lens~\cite{rao_table_1994}, LineUp~\cite{gratzl_lineup:_2013}, and Taggle~\cite{furmanove_taggle_2019}).

\section{InstanceFlow}

The InstanceFlow interface consists of two main components, as illustrated in Figure~\ref{fig:teaser}: The \emph{Flow View}~\ref{fig_label:teaser_1} shows a Sankey diagram of the model's instance predictions throughout the selected training epochs; the \emph{Tabular View}~\ref{fig_label:teaser_3} lists detailed temporal instance information including performance scores~\ref{fig_label:teaser_4}. 

The Flow View supports different levels of granularity.
In its basic form, the Flow View visualizes \enquote{class changers} in a Sankey diagram.
\emph{Distribution Bar Charts} emphasize the fraction of correctly versus incorrectly classified instances.
At the finest granularity, \emph{Instance Glyphs} encode each individual sample, with \emph{Instance Traces} connecting the instances to reveal their \added{classification history, i.e. their} \enquote{paths} through the epochs. 

The Tabular View lists all instances along with their associated predictions over time and allows finding, ranking, and grouping instances via custom instance-level performance measures.

The Flow View and Tabular View are fully linked, such that traced or selected instances are highlighted in both views simultaneously.

\begin{figure}[b!]
  \centering
  \scriptsize\sffamily
  \begin{tabular}{@{}c@{\hspace{.04\columnwidth}}c@{}}
    \includegraphics[width=.475\columnwidth]{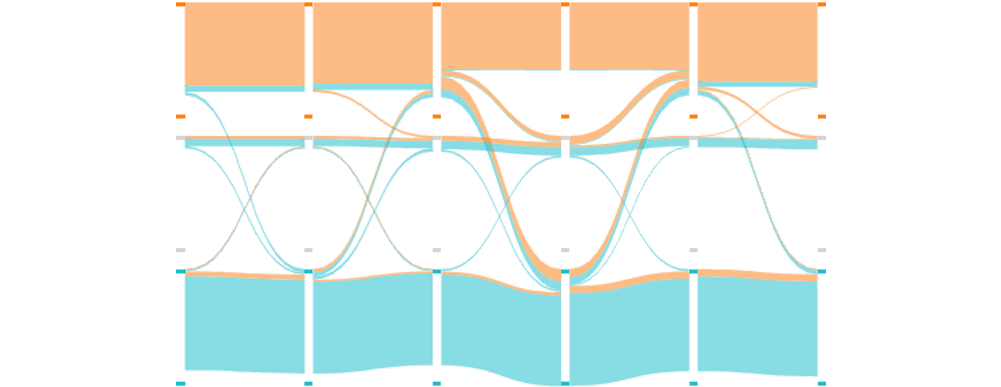} &
    \includegraphics[width=.475\columnwidth]{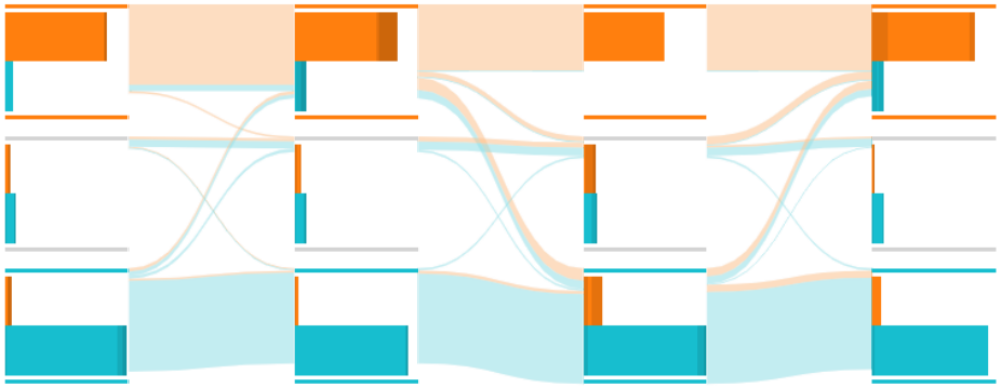} \\
    Flow View (basic)\phantomsection\label{subfig:flow-dense} &
       Distribution Bar Charts\phantomsection \label{subfig:flow-dist}\\[1ex]
    \includegraphics[width=.475\columnwidth]{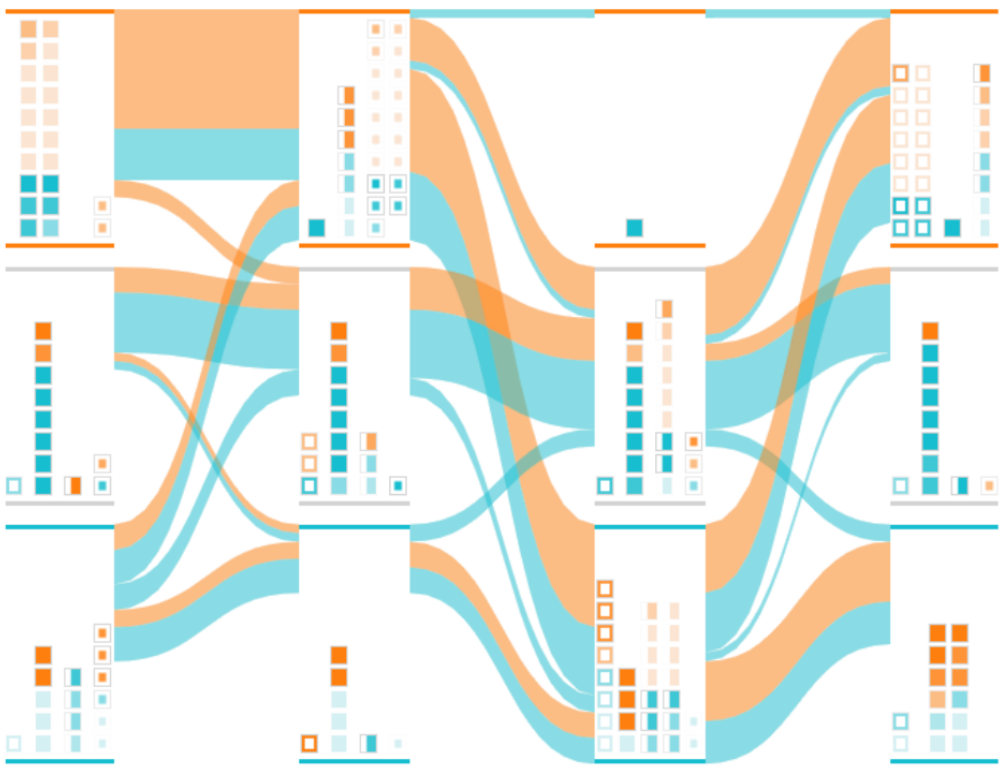} &
    \includegraphics[width=.475\columnwidth]{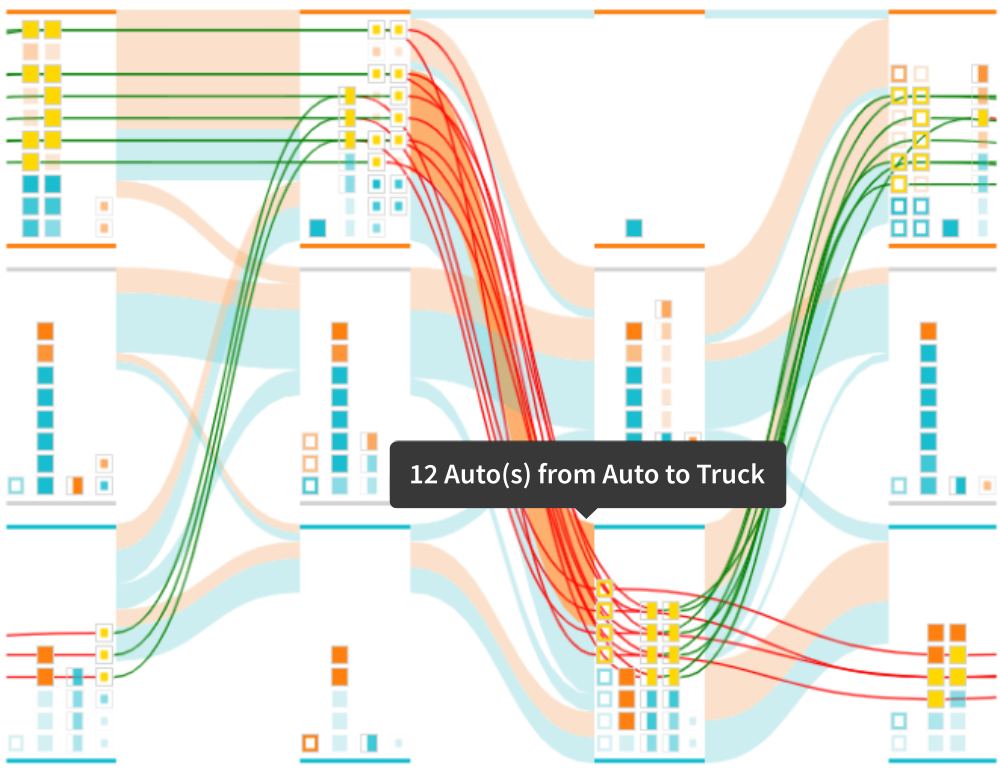} \\
    Instance Glyphs\phantomsection\label{subfig:flow-glyph} &
        Instance Traces\phantomsection\label{subfig:flow-trace}
  \end{tabular}
  \caption{Flow View with all possible extensions. Distribution Bar Charts emphasize the class distribution, Instance Glyphs show the underlying instances, and Instance Traces reveal individual paths through epochs.}
  \label{fig:flow-view}
\end{figure}

\subsection{Flow View}

\added{To visualize the overall flow of instances between classes, we use the well-established Sankey diagram.}
The flow visualization can be seen in Figure~\ref{fig:flow-view} (\hyperref[subfig:flow-dense]{top left}), where the \(x\)-axis denotes the epoch and the \(y\)-axis denotes the predicted classes.
\removed{A~Sankey diagram visualizes the Flow, i.e.,}\added{The thickness of each band in the diagram encodes} how many instances move from one class to another in the following epoch.
The user selects classes of interest, and each class is assigned to a vertical region in the Sankey diagram.
All non-selected classes are aggregated as \enquote{Other} and also assigned to a dedicated vertical region.
The range of epochs to be visualized can be selected via an epoch slider.
Hovering over a section of the Sankey diagram reveals the exact number of instances moving between the corresponding classes.
Clicking on a section of the Sankey diagram selects those instances.

\paragraph{Distribution Charts}

\added{To emphasize the class distributions at each epoch, horizontal Distribution Bar Charts, placed between the Flow visualizations, can be switched on (see Figure~\ref{fig:flow-view}, \hyperref[subfig:flow-dist]{top right}).
While this information is already implicitly encoded by the thicknesses at the borders of each band in the Flow View, the horizontal bar charts facilitate a quantitative comparison.}
\removed{The height of each flow implicitly encodes the distribution of the predictions in each class.
To emphasize the distributions at each epoch, they can be explicitly encoded in Distribution Bar Charts placed between the Flow visualizations (see Figure~\ref{fig:flow-view}, \hyperref[subfig:flow-dist]{top right}).}

\paragraph{Instance Glyphs}

For a more detailed view, the \added{individual} instances \removed{themselves }can be represented by rectangular glyphs (see Figure~\removed{\ref{subfig:flow-glyph}}\added{\ref{fig:flow-view}}, \hyperref[subfig:flow-glyph]{bottom left}).
The color of the Instance Glyphs denotes the actual class of the instances (e.g., \AUTOBOX{} and \TRUCKBOX{} in Figure~\ref{fig:flow-view}).
\removed{Additional information is encoded in their shape, opacity, and position.}
The shape \removed{and horizontal position }indicate\added{s} if the instance predictions are temporally stable~(\STABLEBOX), \removed{changing}\added{coming} from a different class~(\INBOX), leaving for a different class~(\OUTBOX), or coming from and leaving for different classes~(\INOUTBOX).
\added{The horizontal positions encode the same information, with glyphs for incoming instances placed at the left, outgoing ones at the right, and stable ones at the center.
To visually rank the instances by their \enquote{importance}, t}\removed{T}he opacity and vertical \removed{box} position \added{of each glyph} encode one of the calculated numerical difficulty measures described in Section~\ref{sec:metrics}\removed{, which visually ranks the instances by the model's performance}.

\paragraph{Instance Traces}

To allow users to track the path of specific instances throughout the training epochs, their traces can be visualized as lines connecting the corresponding instance glyphs (see Figure~\removed{\ref{subfig:flow-trace}}\added{\ref{fig:flow-view}}, \hyperref[subfig:flow-trace]{bottom right}).
The color of these Instance Traces indicates if the instance is moving to the correct (\textcolor{correct_line}{\rule[.4ex]{1em}{1.6pt}}) or incorrect (\textcolor{incorrect_line}{\rule[.4ex]{1em}{1.6pt}}) class.
Instance Traces are only shown for selected instances, i.e., by clicking on an Instance Glyph\removed{,}\added{or} a section of the Sankey diagram, or \added{by} selecting instances from the Tabular View.

\subsection{Tabular View}
\label{Notation}

The per-class distribution flow is effective for finding anomalies in the learning process, but recognizing specific instances can be hard due to the high information density.
To facilitate the tasks of identifying problematic instances~(\ref{task:inst-diff}--\ref{task:inst-oscil}), all instances are organized in a sortable, filterable, and flexibly customizable table.
The LineUp technique allows an interactive exploration of rankings based on multiple attributes of a given tabular dataset~\cite{gratzl_lineup:_2013}.
Each instance is a row in the LineUp table.
By default, only instances with at least one incorrect classification are shown in InstanceFlow's Tabular View.
The columns include the input data (i.e., images in case of image classification), the ground truth class label, and several \enquote{difficulty} measures defined in Section~\ref{sec:metrics}.
One column shows the class predictions over time as a colored heatmap (see sixth column in Figure~\ref{fig:teaser}), using a categorical color scheme to encode the sequence of predicted classes.
An additional column shows a histogram of correct~(\textcolor{correct_bar}{$\blacksquare$}), incorrect~(\textcolor{incorrect_bar}{$\blacksquare$}), and other~(\textcolor{other_bar}{$\blacksquare$}) predictions (see fifth column in Figure~\ref{fig:teaser}).
Here, \enquote{incorrect} refers to predictions of wrong classes from among the selected subset of classes, while \enquote{other} refers to wrong predictions of non-selected classes.
The encodings in both of these columns can be switched between time-dependent heatmaps and summarizing histograms.

\begin{figure}
  \centering
    \includegraphics[width=\columnwidth,tics=10]{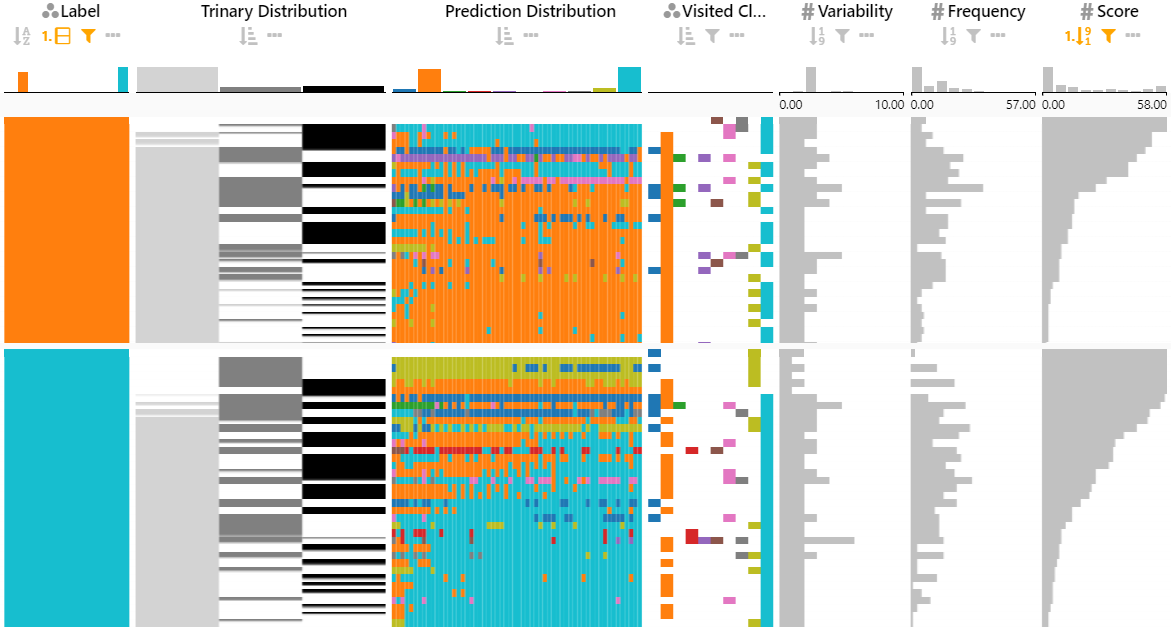}
  \caption{Condensed mode of all instances revealing patterns of successful learning in the classification process.}
  \label{fig:lineup_full_overview}
\end{figure}
\begin{figure}
  \centering
    \begin{overpic}[width=\columnwidth,tics=10]{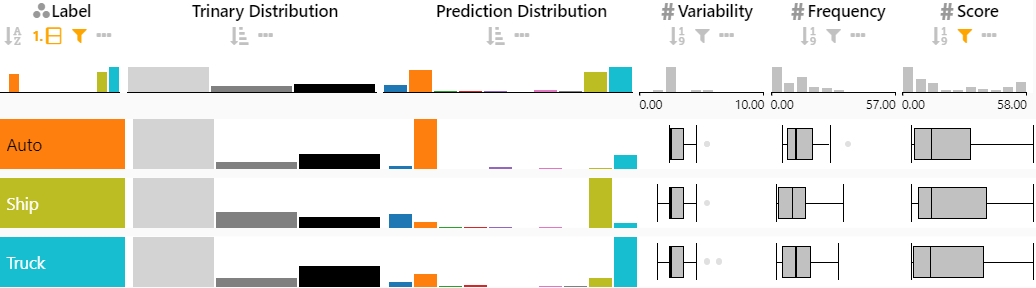}
    \end{overpic}
  \caption{Summary mode of the Tabular View. The overview is similar to a confusion matrix, with correct classifications along the diagonal.}
  \label{fig:lineup_full_overview_summary}
\end{figure}

The LineUp technique includes a number of interactive features to explore the instance predictions:
(1)~\emph{Ranking:} instances can be sorted by each of the attributes in the columns, or by user-defined combinations of attributes;
(2)~\emph{Filtering:} Users can further filter the instances, again either by an individual column's value\removed{,} or by using combined filters on multiple columns.
Advanced filtering with respect to temporally changing attributes is possible via regular expressions.
(3)~\emph{Grouping and Aggregating:} Users can gain an overview of the table by switching to a display mode in which the height of each row is reduced to a minimal height of a single pixel (see Figure~\ref{fig:lineup_full_overview}).
As a result, the previously individual heatmaps and bar charts now form a dense, two-dimensional table that reveals overall patterns, similar to the Table Lens technique~\cite{rao_table_1994}.
User\added{s} can further condense the display by using the group aggregation feature of LineUp, which shows only summary visualizations for the selected classes.
Depending on the attribute type, classes are summarized using histograms or box plots (see Figure~\ref{fig:lineup_full_overview_summary}).
The summary histograms for the prediction distributions encode the same information as a confusion matrix.

\paragraph{Difficulty Measures}
\label{sec:metrics}

The ranking and filtering operations can help users to identify interesting instances when used in conjunction with measures that describe how difficult an instance is to classify.
In this section, we describe three such measures.

Let \(m\) be the total number of instances, \(n\) the number of classes, and \(k\) the number of selected epochs.
Let \(C(i)\) be the actual class of instance \(i\) and \(P(i, j)\) the prediction for instance \(i\) in epoch \(j\).

The \emph{misclassification score} \(S\) of an instance is the fraction of epochs in which it was assigned to the wrong class:  \( S(i) = (1/k)\sum_{j=1}^{k} [P(i,j) \neq C(i)] \).
A\removed{n} misclassification score of~\(0\) means the model predicted the correct class in every epoch, whereas a score of~\(1\) means that the model never predicted the correct class.

The \emph{variability} \(V\) is the ratio of how many classes were predicted for an instance across all epochs: \( V(i) = (1/n) \lvert\{P(i,j)\}_{j\in\{1,\ldots,k\}}\rvert \).
A~variability of~\(1/n\) means that the model predicted the same class in every epoch, whereas a variability of~\(1\) means that the model predicted every possible class at least once.

The \emph{frequency} \(F\) is the ratio of epoch transitions for which the model's prediction jumps between classes: \( F(i) = 1/(k-1){\sum_{j=1}^{k-1} [P(i,j) \neq P(i,j+1)]} \).
A~frequency of \(0\) means that an instance always stayed in the same class, whereas a frequency of \(1\) means that the prediction changed after every epoch.

\subsection{Relationship between Views and Tasks}

The different levels of detail in the Flow View and the Tabular View with its different numerical measures have complementary strengths.
Table~\ref{tab:tasks} assigns the proposed user tasks from Table~\ref{tab:proposed_tasks} to the different visualizations\slash measures, depending on whether the tasks are well supported~ (\yes), partially supported~(\meh), or not supported.
Instance-focused tasks~(\ref{task:inst-diff}--\ref{task:inst-oscil}) are primarily enabled by the Flow View at full detail, whereas the epoch-focused tasks~(\ref{task:epoch-dist}--\ref{task:epoch-move}) are better supported by the more aggregated visualizations.
The Tabular View supports a wide range of tasks.

For the instance-level analysis, the Flow View is focused primarily on the free exploration of a classifier's behavior, or for tracing individual instances once they have been located.
This location of interesting instances is enabled by the Tabular View with its ranking and filtering operations based on the difficulty measures.
For epoch-level analysis, the aggregated Sankey visualization provides a good overview of the class distributions and overall flows.

\begin{table}[th]
    \centering
    \caption{Comparison of InstanceFlow visualization components \& difficulty measures with respect to the user tasks introduced in Section~\ref{sec:tasks}.}
    \label{tab:tasks}
    \footnotesize\sffamily
    \begin{tabular}{l@{\quad}c*{3}{@{~~}c}@{\quad~}c*{2}{@{~~}c}}
        \toprule
        \textbf{Visualization\,\slash\,Metric} & 
        \ref{task:inst-diff} & \ref{task:inst-path} & \ref{task:inst-visit} & \ref{task:inst-oscil} & \ref{task:epoch-dist} & \ref{task:epoch-wrong} & \ref{task:epoch-move} \\
        \midrule
        Flow View (basic)       &      &      &      &      & \meh & \yes & \yes \\
        Distribution Bar Charts &      &      &      &      & \yes & \yes & \yes \\
        Instance Glyphs \& Traces&\meh & \yes & \meh & \meh &      & \meh & \yes \\
        Tabular View            & \yes & \meh & \yes & \yes & \yes & \yes & \yes \\
        \midrule
        Misclassification Score & \yes &      & \meh & \meh &      &      &      \\
        Variability             & \meh &      & \yes & \meh &      &      &      \\
        Frequency               & \meh &      & \meh & \yes &      &      &      \\
        \bottomrule
    \end{tabular}
\end{table}

\subsection{Implementation}

InstanceFlow is a client-side web application built using the React framework.
The code for InstanceFlow is available on GitHub\footnote{Repository: \url{https://github.com/puehringer/InstanceFlow}}.
A~deployed prototype of InstanceFlow with example datasets and the ability to upload new datasets is available online\footnote{Prototype: \url{https://instanceflow.pueh.xyz/}}.

\added{\section{Usage Scenario: CIFAR-10 Image Classification}}

For this \removed{use case}\added{usage scenario}, we consider a simple neural network trained to classify \removed{thumbnail }images from the CIFAR-10 dataset.
This training and test set is a popular choice in the machine learning field and consists of 60,000 color images (\(32 \times 32\)~px) divided into 10 different classes such as \emph{Auto}, \emph{Truck}, \emph{Cat}, and \emph{Dog}~\cite{Krizhevsky2009LearningML}.
A~model developer buil\removed{d}\added{t} a simple CNN using Keras~\cite{KerasExample}.
The developer is satisfied with the overall performance, but notices errors for \textcolor{auto_color!100}{\emph{Auto}} and \textcolor{truck_color!100}{\emph{Truck}} instances.
The model developer (user) analyzes the training process with InstanceFlow to better understand what causes these errors.

\begin{figure}
  \centering
  \begin{overpic}[width=\columnwidth,tics=10]{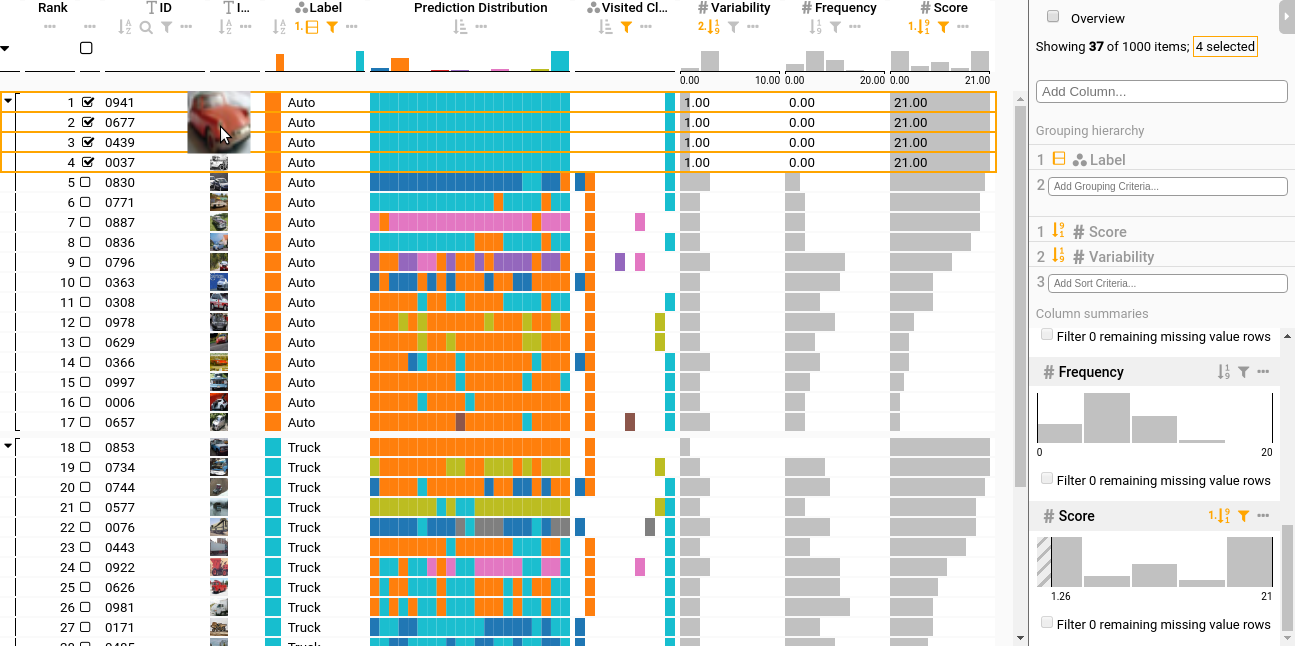}
  \end{overpic}
  \caption{InstanceFlow showing \textcolor{auto_color!100}{\emph{Auto}} and \textcolor{truck_color!100}{\emph{Truck}} instances of CIFAR-10 sorted by high \added{misclassification} score and low variability, and grouped by the ground truth label.}
  \label{fig:app_state_04}
\end{figure}

\begin{figure}
  \centering
  \begin{overpic}[width=\columnwidth,tics=10]{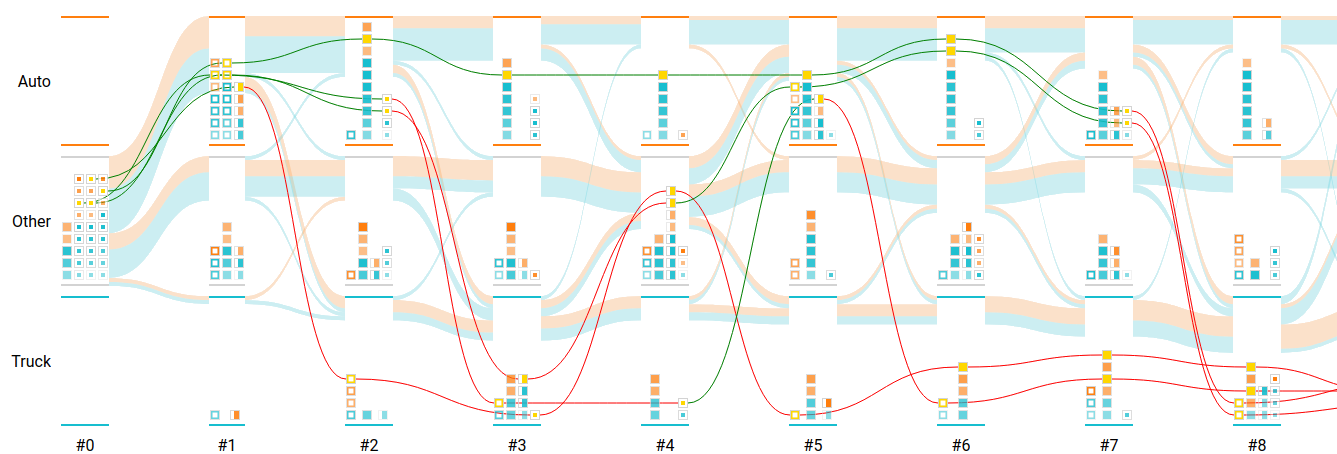}
  \end{overpic}
  \caption{Instance Traces for several selected \textcolor{auto_color!100}{\emph{Auto}} images of bulky, antique cars. These images are correctly classified as \textcolor{auto_color!100}{\emph{Auto}} in the beginning, but tend to be consistently classified as \textcolor{truck_color!100}{\emph{Truck}} over time.}
  \label{fig:app_state_05}
\end{figure}

\begin{enumerate}[leftmargin=*, itemsep=.25ex]
    \item The user trains the neural network to classify CIFAR-10 images, and loads the classification results into InstanceFlow.
    \item In the Tabular View, the user groups instances by their actual class and enables the condensed mode with the predictions shown as histograms.
    This gives the user a hint about class confusion over the total selected epoch range\removed{ (i.e., similar to the time-integral of a confusion matrix)}.
    \item The user notices that most classes are predicted correctly (with the bin for correct classification being by far the highest).
    However, for the \textcolor{auto_color!100}{\emph{Auto}} class the user finds that the \textcolor{truck_color!100}{\emph{Truck}} bar is similarly high as the actual class (and vice versa).
    This is an immediate hint for a high class confusion between \textcolor{auto_color!100}{\emph{Auto}} and \textcolor{truck_color!100}{\emph{Truck}}. 
    \item The user is now interested in why the neural network incorrectly classifies  \textcolor{auto_color!100}{\emph{Auto}} images as \textcolor{truck_color!100}{\emph{Truck}}.
    To focus on this confusion, the user hides all other classes.
    Additionally, the user filters the instances to only show those classified as \textcolor{auto_color!100}{\emph{Auto}} or \textcolor{truck_color!100}{\emph{Truck}} at least once.
    Finally, the user switches from the condensed mode to the normal mode to gain access to the actual underlying instances.
    Sorting by high misclassification score and low variability reveals to the user that the most problematic instances are mainly \textcolor{auto_color!100}{\emph{Auto}} images classified as \textcolor{truck_color!100}{\emph{Truck}}. 
    \item \begin{minipage}[t]{.58\columnwidth}
        Now the user notices a common pattern: the topmost images all show old and bulky cars (see  Figure~\ref{fig:app_state_04}).
        \end{minipage}%
        \begin{minipage}[t]{.36\columnwidth}
            \vspace*{-5pt}\hspace*{\fill}%
            \includegraphics[height=24px]{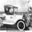}~%
            \includegraphics[height=24px]{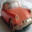}~%
            \includegraphics[height=24px]{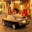}%
        \end{minipage}
    \item The user proceeds by investigating the flow of these images (see Figure~\ref{fig:app_state_05}). 
    It becomes clear that all of them were correctly classified in early epochs, but then suddenly change to \textcolor{truck_color!100}{\emph{Truck}} one after another.
    \item The user checks the traces of random modern-looking cars and finds, in stark contrast to the previous instances, that many of them are temporally stable and correctly classified \emph{after an initial misclassification}.
    This leads the user to hypothesize that the network, over time, learns features that tend to prioritize modern cars over bulky, antique cars.
    \item The user can use these new insights in the subsequent model development or refinement process, e.g., by increasing the number of problematic instances in an attempt to improve the accuracy and temporal stability for \textcolor{auto_color!100}{\emph{Auto}} images.
\end{enumerate}

\section{Limitations \& Future Work}




\paragraph{Scalability}

Due to the combination of aggregated information in the basic Flow View (without Instance Glyphs and Traces) with the functionality of the Tabular View, InstanceFlow scales well to large datasets.
However, for more than \(\sim100\) \emph{selected} instances and at full detail, the InstanceFlow visualization can get cluttered.
Additionally, with each selected class, the number of possible paths in the Sankey visualization increases.
Thus, additional class aggregation or automatic class and/or instance selection mechanisms would be necessary for exploring datasets with many (\(\gtrsim 15\)) classes.

\paragraph{Comparison of Datasets}

A~comparison of multiple classification models can be helpful for evaluating the effectiveness of modifications applied during model development.
\removed{A~}\added{We introduced a} combined temporal-comparative approach \removed{was introduced }for class-level analysis with ConfusionFlow\cite{confusionflow}.
However, it is not straightforward how to extend InstanceFlow to allow similar comparison tasks in a\added{n} \removed{way that is more }effective \added{way}\removed{than simply using two InstanceFlow visualizations side by side}.

\section{Conclusion}

We introduced InstanceFlow, a visualization of the evolution of instance classifications in machine learning.
The Flow View supports users in understanding the temporal progression of predicted class distributions \removed{in}\added{using} a Sankey diagram.
Detailed visualizations allow\removed{s} users to trace the predictions for individual instances over time.
Interesting instances can be located effectively in the Tabular View, which allows ranking and filtering by numerical difficulty measures.
With its different aggregation levels, InstanceFlow bridges the gap between class-level and instance-level performance evaluation while enabling a full temporal analysis of the training process.



\bibliographystyle{abbrv-doi-hyperref}

\bibliography{sources}
\end{document}